\documentclass[]{ceurart}
\usepackage{float}
\usepackage{graphicx}

\begin{document}
\copyrightyear{2023}
\copyrightclause{Copyright for this paper by its authors.
  Use permitted under Creative Commons License Attribution 4.0
  International (CC BY 4.0).}

\conference{CLEF 2023 -- Conference and Labs of the Evaluation Forum, 
	September 18--21, 2023, Thessaloniki, Greece}

\title{ARC-NLP at PAN 2023: Hierarchical Long Text Classification for Trigger Detection}

\author[1]{Umitcan Sahin}[%
orcid=0000-0001-9594-3148,
email=ucsahin@aselsan.com.tr,
]
\address[1]{Aselsan Research Center, 06378, Ankara, Turkey.}

\author[1]{Izzet Emre Kucukkaya}[%
orcid=0009-0006-2877-8713,
email=ekucukkaya@aselsan.com.tr,
]

\author[1]{Cagri Toraman}[%
orcid=0000-0001-6976-3258,
email=ctoraman@aselsan.com.tr,
]

\begin{abstract}
Fanfiction, a popular form of creative writing set within established fictional universes, has gained a substantial online following. However, ensuring the well-being and safety of participants has become a critical concern in this community. The detection of triggering content, material that may cause emotional distress or trauma to readers, poses a significant challenge. In this paper, we describe our approach for the Trigger Detection shared task at PAN CLEF 2023, where we want to detect multiple triggering content in a given Fanfiction document. For this, we build a hierarchical model that uses recurrence over Transformer-based language models. In our approach, we first split long documents into smaller sized segments and use them to fine-tune a Transformer model. Then, we extract feature embeddings from the fine-tuned Transformer model, which are used as input in the training of multiple LSTM models for trigger detection in a multi-label setting. Our model achieves an F1-macro score of 0.372 and F1-micro score of 0.736 on the validation set, which are higher than the baseline results shared at PAN CLEF 2023.  
\end{abstract}


\maketitle

\section{Introduction}
Fanfiction has been incredibly popular in recent years, especially among online communities. It is a lively and imaginative form of literary expression. It entails the development of fresh storylines, characters, and situations based on pre-existing fictional worlds, giving fans the chance to develop their favorite stories and pursue original, inventive concepts. As the Fanfiction community expands, there is an increasing need to address crucial issues related to participant safety and well-being.

The existence of triggering content is a major issue in the world of Fanfiction. Content that can cause extreme negative emotional reactions or the traumatization of people is referred to as triggering content. It might touch on issues like abuse, violence, mental health, or other upsetting topics. It is essential to create systems for properly identifying and managing triggering content given the diverse and frequently fragile character of Fanfiction readers and writers.

In this study, as Aselsan Research Center - Natural Language Processing team (ARC-NLP), we propose a method for trigger detection in long text documents using natural language processing (NLP) and machine learning techniques. We seek to train a classification algorithm capable of precisely recognizing multiple triggering contents by using the concept of hierarchical recurrence over Transformer-based language models \cite{Pappagari:2019, Dai:2022, Park:2022}. In our method, we first split long Fanfiction documents into smaller sized segments with an overlap between each consecutive segment. We use these segments to fine-tune a Transformer-based language model. Then, we extract each segment's feature embedding from the fine-tuned Transformer model, which are used in the training of multiple LSTM models. Finally, we combine the predictions of the trained LSTM models to generate trigger labels for multi-class and multi-label classification. To the best of our knowledge, we are first to use the techniques and methods described in this paper in the context of trigger detection in Fanfiction.   

\section{Task Description}
In the context of trigger detection at PAN CLEF 2023 \cite{Bevendorff:2023, Wiegmann:2023b}, our objective is to assign warning labels to Fanfiction documents that may contain content capable of causing discomfort or distress (known as triggering content) \cite{Wiegmann:2023b, Wolska:2022, Wiegmann:2023}. Specifically, trigger detection is posed as a multi-label document classification task, aiming to assign the appropriate trigger warnings to each document without exceeding the necessary labels. It is important to note that all trigger warnings are determined from the perspective of the document's author, meaning that the author decides which specific trigger(s) the document contains. There are 32 distinct trigger labels including \textit{pornographic-content}, \textit{violence}, \textit{death} and \textit{sexual-assault} etc\footnote{https://pan.webis.de/clef23/pan23-web/trigger-detection.html}. Each document can contain more than one trigger label, which leads to the need to adopt a multi-class and multi-label approach.  

\section{Dataset}
\begin{table}[ht]
   \caption{Trigger detection dataset shared at PAN CLEF 2023. The number of documents, average number of words in each document and presence of labels for \textit{training}, \textit{validation}, and \textit{test} sets are given.}
   \label{tab:dataset}
   \begin{tabular}{lrrc}
     \toprule
     &\#Document&\#Avg. Words&Labels\\
     \midrule
     Train & 307,102 & 2,350 & \checkmark \\
     Validation & 17,104 & 2,336 & \checkmark \\
     Test & 17,040 & 2,338 & - \\
   \bottomrule
 \end{tabular}
\end{table}

The dataset for trigger detection at PAN CLEF 2023 comprises of fanfiction pieces sourced from archiveofourown.org (AO3) \cite{Wolska:2022, Wiegmann:2023}. Each document falls within the range of 50 to 6,000 words and is accompanied by one or more trigger labels. Table \ref{tab:dataset} shows the number of documents, average number of words, and presence of labels for training, validation, and test sets. As mentioned before, the label set encompasses 32 distinct trigger warnings, exhibiting a distribution where certain labels are frequently encountered while the majority of labels are relatively uncommon. Table \ref{tab:label_dist} shows the class distribution ratios with respect to 32 trigger labels in the training and validation sets. As seen, the classes are greatly imbalanced. Furthermore, we also note that the class distributions between the training and validation sets are very similar to each other, which suggests that they come from the same distribution.  

\begin{table}[ht]
\caption{Class distribution ratios with respect to 32 trigger labels in the training and validation sets.}
\label{tab:label_dist}
\centering
\begin{tabular}{rlrr}
\toprule
& Class & Train & Validation  \\
\midrule
1 & \textit{pornographic} & 77.52\% & 77.33\%  \\
2 & \textit{violence} & 9.48\% & 9.46\% \\
3 & \textit{death} & 6.77\%  & 6.75\%  \\
4 & \textit{sexual-assault} & 10.20\% & 10.18\%   \\
5 & \textit{abuse} & 7.22\% & 7.21\%  \\
6 & \textit{blood} &  4.92\% & 4.90\%  \\
7 & \textit{suicide} & 2.67\% &  2.67\%   \\
8 & \textit{pregnancy} & 4.44\% & 4.44\%  \\
9 & \textit{child-abuse} & 2.34\% & 2.34\% \\
10 & \textit{incest} &  4.39\% & 4.38\% \\
11 & \textit{underage} & 2.90\% & 2.89\%   \\
12 & \textit{homophobia} & 1.61\% & 1.61\%  \\
13 & \textit{self-harm} & 1.71\% & 1.71\%  \\
14 & \textit{dying} & 2.44\% & 2.44\%  \\
15 & \textit{kidnapping} & 1.46\% & 1.45\%  \\
16 & \textit{mental-illness} & 1.36\% &  1.36\% \\
17 & \textit{dissection} & 0.56\% & 0.55\% \\
18 & \textit{eating-disorder} & 0.39\% & 0.40\% \\
19 & \textit{abduction} & 0.35\% &  0.34\% \\
20 & \textit{body-hatred} & 0.44\% & 0.44\%  \\
21 & \textit{childbirth} & 0.28\% &  0.28\% \\
22 & \textit{racism} & 0.13\% & 0.13\%   \\
23 & \textit{sexism} & 0.17\%  &  0.17\% \\
24 & \textit{miscarriage} & 0.16\% &  0.17\% \\
25 & \textit{transphobia} & 0.12\% & 0.12\% \\
26 & \textit{abortion} & 0.11\% & 0.12\% \\
27 & \textit{fat-phobia} & 0.24\% & 0.24\% \\
28 & \textit{animal-death} & 0.06\% & 0.07\% \\
29 & \textit{ableism}  & 0.09\% & 0.09\%  \\
30 & \textit{classism} & 0.06\% & 0.06\% \\
31 & \textit{misogyny} & 0.07\% & 0.08\% \\
32 & \textit{animal-cruelty} & 0.05\% & 0.05\%  \\
\hline
\end{tabular} %
\end{table}

\section{Proposed Method: Hierarchical Recurrence over Transformer-based Language Model}
\label{sec:proposed_method}
In this section, we describe our method of hierarchical recurrence over Transformer-based model for long text classification of multi-label trigger detection. The diagram of the model is shown in Figure \ref{fig:model}. As shown in the figure, we divide our methodology into four parts: 1) Segmentation, 2) Tokenization, 3) Feature extraction, and 4) Model training, which are explained in detail below.   
\subsection{Segmentation}
According to the information provided in Table \ref{tab:dataset}, the average word count in each document within the training set exceeds 2000. Traditional Transformer-based language models such as BERT \cite{Devlin:2018} employ tokenizers with a maximum length of 512 tokens and typically truncate the remaining text. However, this approach is inadequate for accurately classifying long documents since crucial information may be omitted through truncation, resulting in subpar classification performance. Therefore, we follow a similar approach to \cite{Pappagari:2019, Dai:2022, Park:2022, Ozcelik:22} in our segmentation method. We first apply text processing to the documents by 
\begin{itemize}
    \item removing HTML tags,
    \item removing URLs,
    \item removing English stop words \cite{Bird:2009}, and
    \item lower-casing all text.
\end{itemize}
Then, we split the processed document into segments (i.e., text chunks) of 200 words (i.e., ${w_0,...,w_{200}}$) with an overlap of 50 words between each consecutive segment as shown in Figure \ref{fig:model}. We assign the original document label to each segment. In other words, we represent each document in the training set with a variable-length sequence of 200-word segments where each segment is assigned the original document label.  

\subsection{Tokenization}
\begin{figure}[p]
  \centering
  \includegraphics[width=\linewidth]{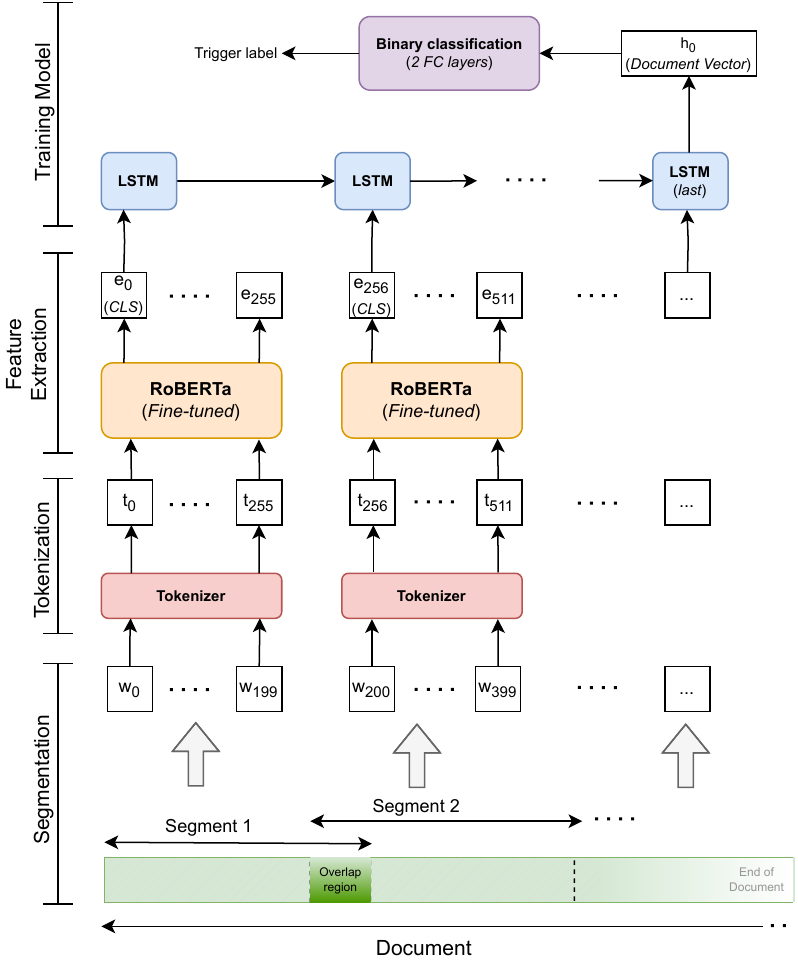}
  \caption{Hierarchical recurrence over Transformer-based model diagram for trigger detection. Each document is split into $200$-word segments with a $50$-word overlap between the consecutive segments. ROBERTa \cite{Liu:19} along with its tokenizer (256 tokens) is used as the Transformer model and fine-tuned using all the segmented documents in the training set. Transformer model and its tokenizer are shared throughout the network. CLS embeddings are extracted for each segment of a given document, which are then used as input feature vectors for the training of the LSTM model. For the classifier, two fully-connected linear layers with ReLU activation and binary cross-entropy (BCE) loss are used.}
  \label{fig:model}
\end{figure}

For tokenization, we fine-tune a Transformer-based RoBERTa model \cite{Liu:19} using the sequence of segments obtained after our segmentation method. We use the base-version of the model at HuggingFace\footnote{https://huggingface.co/roberta-base} with a learning rate of $1e-5$, epoch number $3$, and training batch size of $8$. We also use the corresponding RoBERTa model's tokenizer with a maximum length of $256$ tokens. After the tokenization step, we represent each $200$-word segment by the corresponding tokens with size $256$ (i.e., ${t_0,...,t_{255}}$ in Figure \ref{fig:model}). 

\subsection{Feature Extraction}
For this method, we feed forward the segment tokens obtained after our tokenization method to the fine-tuned RoBERTa model and construct segment embeddings. We extract the embedding of the CLS token (i.e., a special classification token used for classification tasks) from the last hidden layer of the fine-tuned RoBERTa model for each segment and use them as our feature vectors, which are input to the LSTM model.   

\subsection{Model Training}
After feature extraction, we train a single layer LSTM network with a hidden unit size of 100 and a batch size of 8. We use Pytorch's implementation of the Stochastic Gradient Descent (SGD) optimizer with learning rate of $0.01$ \cite{Paszke:17}. In the classification step, we use two fully connected linear layers with sizes $100$ and ReLU \cite{Agarap:18} as the activation function between them. Furthermore, we broke down the multi-label trigger detection problem into a multiple binary classification problem. Therefore, we train the aforementioned LSTM model for each trigger class (32 in total) in a one-vs-all classification setting. Finally, we use Binary Cross Entropy (BCE) as our loss function. 

As seen in Table \ref{tab:label_dist}, the trigger classes are greatly imbalanced. For instance, while the $77.52\%$ of the training data includes \textit{pornographic} content, only the $0.05\%$ of it consists of \textit{animal cruelty}. There are many methods such as oversampling the underrepresented classes, that are proposed to solve the class imbalance problem for deep neural networks \cite{Buda:18}. In this study, we solve this problem by changing the weights of the underrepresented classes during the back-propagation updates. We do this by incorporating a weight of positive examples for the loss function. For example, if a dataset contains 20 positive and 1000 negative examples of a single class, then weight for the positive class is assigned as $1000/20=50$ in the loss function. In this way, the loss would act as if the dataset contains $20\times 50=1000$ positive examples. In our final model, we assign positive class weights to the loss function for the trigger classes from 15 (i.e., \textit{kidnapping}) to 32 (i.e., \textit{animal-cruelty}). For the first 14 trigger classes, we do not assign any positive class weights to BCE loss function.  

Finally, for each trigger class, we train the LSTM model up to 10 epochs and save the best performing model with respect to the highest F1 scores achieved on the validation set. We then combine the predictions of the 32 trained LSTM models to produce our final multi-label trigger labels.  

\section{Experiments}
\label{sec:experiments}
In this section, we describe the baseline methods used for multi-label trigger detection in Fanfiction, and share the classification performance of our model on the validation set.
\subsection{Baselines}
\begin{itemize}
    \item \textbf{BERT}: We use HuggingFace's implementation of the Transformer-based BERT model\footnote{https://huggingface.co/bert-base-uncased} and fine-tune this model with the training set for trigger detection. We set learning rate to $1e-5$, number of training epochs to $5$, and batch size to $8$. We use multi-label classification layer with $32$ classes on top of the BERT model. We use the corresponding BERT tokenizer with maximum length of 512 tokens, truncating the rest of the document in the training set.
    \item \textbf{RoBERTa-Segment}: We use the same RoBERTa model that is fine-tuned with the segmented tokens described in Section \ref{sec:proposed_method}.
    \item \textbf{TFIDF+XGBoost}: This baseline is shared by the organizers at PAN CLEF 2023 for trigger detection \cite{Wiegmann:2023}\footnote{https://github.com/pan-webis-de/pan-code/tree/master/clef23/trigger-detection/baselines}. It uses Gradient Boosted Trees \cite{Chen:2016} based on a TF-IDF \cite{Salton:84} document vectors.
\end{itemize}

\subsection{Validation Results}
\begin{table}
\caption{F1-macro and F1-micro scores of our model and baseline methods. The scores are computed on the validation set provided by PAN CLEF 2023.}
\label{tab:validation_results}
\centering
\begin{tabular}{l r r r}
\toprule
Model & F1-macro & F1-micro \\
\midrule  
BERT & 0.0471 & 0.4607 \\
RoBERTa-Segment & 0.1869 & 0.6958 \\
TF-IDF + XGBoost (task baseline) & 0.2575 & 0.7274 \\
Hierarchical LSTM + RoBERTa (ours) & \textbf{0.3720} & \textbf{0.7360}  \\
\hline
\end{tabular}
\end{table}

Multi-label F1-macro and F1-micro scores serve as the primary performance metrics for trigger detection at PAN CLEF 2023\footnote{https://pan.webis.de/clef23/pan23-web/trigger-detection.html}. To assess the effectiveness of our model and the baselines, we compute these scores on the validation set, and the results are presented in Table \ref{tab:validation_results}. Notably, our model outperforms all others in terms of multi-label F1-macro and F1-micro scores.  

The limitations of the BERT model become evident as it struggles due to the constrained token size of 512. Truncating long documents reduces its ability to capture triggering content effectively. While our RoBERTa-Segment baseline improves upon BERT, it still falls short in performance since it lacks the crucial hierarchical recurrence concept integrated into its core architecture.

Among the baselines, the TFIDF+XGBoost approach achieves the highest F1-macro and F1-micro scores. Despite the absence of contextual information, XGBoost compensates by leveraging TF-IDF vector representations, enabling comprehensive coverage of tokens throughout the entire document.

These results underscore the existence of ample room for improvement in the non-trivial task of multi-label trigger detection in Fanfiction. It is evident that further improvements are necessary to enhance the effectiveness of trigger detection methods and address the challenges associated with this task.

\subsection{Class-based Validation Results}
Table \ref{tab:binary_results} shows the binary classification performances of the trained models for each trigger class on the validation set. It is worth noting that by incorporating a positive class weight into the loss function for the classes from 15 to 32, where positive instances are particularly scarce, the model's capability to predict positive class instances is significantly enhanced. Consequently, this leads to an anticipated improvement in the overall performance of multi-label classification.

\begin{table}[h]
\caption{Various binary classification scores of the proposed method (32 hierarchical LSTM models over Transformer-based RoBERTa) on the validation set for trigger detection. Pos. Ratio indicates the true positive class ratio in the validation set. Pos. Pred. Ratio indicates the positive class ratio predicted by the proposed method. The classification performance computed in terms of macro and micro F1, Precision (P), and Recall (R) scores. Overall multi-label classification performances are given at the bottom. }
\label{tab:binary_results}
\centering
\resizebox{\columnwidth}{!}{%
\begin{tabular}{l r r r r r r r r}
\toprule
Class & Pos. Ratio & F1-macro & P-macro & R-macro & F1-micro & P-micro & R-micro & Pos. Pred. Ratio \\
\midrule
\textit{pornographic} & 0.773 & 0.901 & 0.906 & 0.897 & 0.932 & 0.932 & 0.932 & 0.781 \\
\textit{violence}  & 0.095 & 0.715 & 0.744 & 0.694 & 0.912 & 0.912 & 0.912 & 0.074 \\
\textit{death}  & 0.068 & 0.781 & 0.802 & 0.763 & 0.948 & 0.948 & 0.948 & 0.058 \\
\textit{sexual-assault}  & 0.102 & 0.748 & 0.784 & 0.722 & 0.918 & 0.918 & 0.918 & 0.078 \\
\textit{abuse}  & 0.072 & 0.727 & 0.768 & 0.699 & 0.936 & 0.936 & 0.936 & 0.052 \\
\textit{blood}  & 0.049 & 0.758 & 0.791 & 0.733 & 0.959 & 0.959 & 0.959 & 0.039 \\
\textit{suicide}  & 0.027 & 0.797 & 0.841 & 0.764 & 0.981 & 0.981 & 0.981 & 0.021 \\
\textit{pregnancy}  & 0.044 & 0.882 & 0.883 & 0.881 & 0.980 & 0.980 & 0.980 & 0.044 \\
\textit{child-abuse}  & 0.023 & 0.726 & 0.751 & 0.705 & 0.977 & 0.977 & 0.977 & 0.019 \\
\textit{incest}  & 0.044 & 0.837 & 0.835 & 0.839 & 0.973 & 0.973 & 0.973 & 0.044 \\
\textit{underage}  & 0.029 & 0.681 & 0.744 & 0.645 & 0.971 & 0.971 & 0.971 & 0.017 \\
\textit{homophobia}  & 0.016 & 0.711 & 0.750 & 0.682 & 0.984 & 0.984 & 0.984 & 0.012 \\
\textit{self-harm}  & 0.017 & 0.795 & 0.871 & 0.745 & 0.989 & 0.989 & 0.989 & 0.011 \\
\textit{dying}  & 0.024 & 0.678 & 0.735 & 0.644 & 0.975 & 0.975 & 0.975 & 0.015 \\
\textit{kidnapping}  & 0.015 & 0.618 & 0.791 & 0.576 & 0.986 & 0.986 & 0.986 & 0.004 \\
\textit{mental-illness}  & 0.014 & 0.598 & 0.566 & 0.787 & 0.942 & 0.942 & 0.942 & 0.062 \\
\textit{dissection}  & 0.006 & 0.583 & 0.552 & 0.771 & 0.971 & 0.971 & 0.971 & 0.029 \\
\textit{eating-disorder}  & 0.004 & 0.756 & 0.700 & 0.858 & 0.995 & 0.995 & 0.995 & 0.007 \\
\textit{abduction}  & 0.003 & 0.576 & 0.548 & 0.697 & 0.985 & 0.985 & 0.985 & 0.014 \\
\textit{body-hatred}  & 0.004 & 0.639 & 0.595 & 0.765 & 0.988 & 0.988 & 0.988 & 0.012 \\
\textit{childbirth}  & 0.003 & 0.683 & 0.621 & 0.882 & 0.993 & 0.993 & 0.993 & 0.009 \\
\textit{racism}  & 0.001 & 0.605 & 0.633 & 0.587 & 0.998 & 0.998 & 0.998 & 0.001 \\
\textit{sexism}  & 0.002 & 0.577 & 0.563 & 0.599 & 0.996 & 0.996 & 0.996 & 0.003 \\
\textit{miscarriage}  & 0.002 & 0.694 & 0.662 & 0.741 & 0.997 & 0.997 & 0.997 & 0.003 \\
\textit{transphobia}  & 0.001 & 0.722 & 0.682 & 0.785 & 0.998 & 0.998 & 0.998 & 0.002 \\
\textit{abortion}  & 0.001 & 0.531 & 0.523 & 0.549 & 0.997 & 0.997 & 0.997 & 0.002 \\
\textit{fat-phobia}  & 0.002 & 0.780 & 0.780 & 0.780 & 0.998 & 0.998 & 0.998 & 0.002 \\
\textit{animal-death}  & 0.001 & 0.558 & 0.545 & 0.583 & 0.998 & 0.998 & 0.998 & 0.001 \\
\textit{ableism}  & 0.001 & 0.535 & 0.538 & 0.533 & 0.999 & 0.999 & 0.999 & 0.001 \\
\textit{classism}  & 0.001 & 0.500 & 0.500 & 0.500 & 0.999 & 0.999 & 0.999 & 0.000 \\
\textit{misogyny}  & 0.001 & 0.501 & 0.503 & 0.604 & 0.976 & 0.976 & 0.976 & 0.024 \\
\textit{animal-cruelty}  & 0.001 & 0.505 & 0.505 & 0.658 & 0.982 & 0.982 & 0.982 & 0.018 \\
\hline
Multi-label & - & \textbf{0.3720} & 0.3920 & 0.4330 & \textbf{0.7360} & 0.7330 & 0.7400 & - \\
\hline
\end{tabular} %
}
\end{table}

\subsection{Test Results}
We submitted our model as a dockerized image to the TIRA system \cite{Froebe:2023}. The test was conducted on a hardware configuration consisting of a single CPU Core, 10GB of RAM, and a single Nvidia GTX 1080 with 8GB. The test completion time was approximately 150 minutes. The final test results of all the participants for PAN CLEF 2023 Trigger Detection are presented in Table \ref{tab:test_results}. \textit{trigger-detection-baseline} is the TFIDF+XGBoost model explained in Section \ref{sec:experiments}. Furthermore, only the submission with the highest F1-macro score was included for teams with multiple submissions. At the end, our team, named \textbf{\textit{pan23-transformers}}, achieved first place in terms of the multi-label F1-macro score and second place in terms of the multi-label F1-micro score in the leaderboard with our hierarchical recurrence over Transformer-based language model.  

\begin{table}
\caption{The leaderboard in terms of the F1-macro and F1-micro test scores of all participants at PAN CLEF 2023 Trigger Detection. Our team's name is \textbf{\textit{pan23-transformers}}. \textit{trigger-detection-baseline} is the TF-IDF+XGBoost model explained in Section \ref{sec:experiments}.}
\label{tab:test_results}
\centering
\begin{tabular}{l r r r}
\toprule
Team & F1-macro & F1-micro \\
\midrule  
\textbf{\textit{pan23-transformers}} (Ours) & \textbf{0.352} & 0.737  \\
pan23-supergirl & 0.350 & \textbf{0.753}  \\
trigger-detection-baseline & 0.301 & 0.689  \\
pan23-jojo-no-kimyou-na-bouken & 0.228 & 0.557  \\
pan23-marvel-cinematic-universe & 0.225 & 0.616  \\
pan23-sherlock & 0.161 & 0.402  \\
pan23-game-of-thrones & 0.048 & 0.625 \\
\hline
\end{tabular}
\end{table}

\section{Conclusion}
This study presents an approach for detecting triggers in Fanfiction by employing natural language processing (NLP) and machine learning techniques. Our objective is to train a classification algorithm capable of accurately identifying multiple instances of triggering content. In our method, we initially break down lengthy Fanfiction documents into smaller segments, ensuring an overlap between consecutive segments. These segments are then used to fine-tune a Transformer-based language model. From this fine-tuned model, we extract feature embeddings for each segment, which serve as inputs for training multiple LSTM models. Subsequently, the predictions of these trained LSTM models are combined to generate trigger labels for multi-class and multi-label classification. We show that our method that is based on hierarchical recurrence over Transform-based model achieves better classification performance than the baselines used for multi-label trigger detection in Fanfiction. Our model ranks first in terms of the multi-label F1-macro score and second in terms of the multi-label F1-micro score on the test set for PAN CLEF 2023 Trigger Detection. 
Furthermore, Our experimental findings strongly indicate that conventional NLP techniques, such as TF-IDF document vectorization and Transformer-based models with standard tokenization limits (typically set at a maximum length of 512 tokens), exhibit limited performance in the context of multi-class and multi-label classification tasks, particularly when dealing with lengthy documents. These techniques often struggle to effectively handle the complexities associated with the simultaneous prediction of multiple trigger labels in scenarios where extensive text is involved. 

\bibliography{sample-ceur}

\begin{thebibliography}{17}
\expandafter\ifx\csname natexlab\endcsname\relax\def\natexlab#1{#1}\fi
\providecommand{\url}[1]{\texttt{#1}}
\providecommand{\href}[2]{#2}
\providecommand{\path}[1]{#1}
\providecommand{\DOIprefix}{doi:}
\providecommand{\ArXivprefix}{arXiv:}
\providecommand{\URLprefix}{URL: }
\providecommand{\Pubmedprefix}{pmid:}
\providecommand{\doi}[1]{\href{http://dx.doi.org/#1}{\path{#1}}}
\providecommand{\Pubmed}[1]{\href{pmid:#1}{\path{#1}}}
\providecommand{\bibinfo}[2]{#2}
\ifx\xfnm\relax \def\xfnm[#1]{\unskip,\space#1}\fi
\bibitem[{Pappagari et~al.(2019)Pappagari, Zelasko, Villalba, Carmiel, and
  Dehak}]{Pappagari:2019}
\bibinfo{author}{R.~Pappagari}, \bibinfo{author}{P.~Zelasko},
  \bibinfo{author}{J.~Villalba}, \bibinfo{author}{Y.~Carmiel},
  \bibinfo{author}{N.~Dehak},
\newblock \bibinfo{title}{Hierarchical transformers for long document
  classification},
\newblock in: \bibinfo{booktitle}{2019 IEEE Automatic Speech Recognition and
  Understanding Workshop (ASRU)}, \bibinfo{organization}{IEEE},
  \bibinfo{year}{2019}, pp. \bibinfo{pages}{838--844}.
\bibitem[{Dai et~al.(2022)Dai, Chalkidis, Darkner, and Elliott}]{Dai:2022}
\bibinfo{author}{X.~Dai}, \bibinfo{author}{I.~Chalkidis},
  \bibinfo{author}{S.~Darkner}, \bibinfo{author}{D.~Elliott},
\newblock \bibinfo{title}{Revisiting transformer-based models for long document
  classification},
\newblock in: \bibinfo{booktitle}{Findings of the Association for Computational
  Linguistics: EMNLP 2022}, \bibinfo{publisher}{Association for Computational
  Linguistics}, \bibinfo{address}{Abu Dhabi, United Arab Emirates},
  \bibinfo{year}{2022}, pp. \bibinfo{pages}{7212--7230}.
\bibitem[{Park et~al.(2022)Park, Vyas, and Shah}]{Park:2022}
\bibinfo{author}{H.~Park}, \bibinfo{author}{Y.~Vyas},
  \bibinfo{author}{K.~Shah},
\newblock \bibinfo{title}{Efficient classification of long documents using
  transformers},
\newblock in: \bibinfo{booktitle}{Proceedings of the 60th Annual Meeting of the
  Association for Computational Linguistics (Volume 2: Short Papers)},
  \bibinfo{publisher}{Association for Computational Linguistics},
  \bibinfo{address}{Dublin, Ireland}, \bibinfo{year}{2022}, pp.
  \bibinfo{pages}{702--709}. \DOIprefix\doi{10.18653/v1/2022.acl-short.79}.
\bibitem[{Bevendorff et~al.(2023)Bevendorff, Borrego-Obrador, Chinea-R{\'i}os,
  Franco-Salvador, Fr{\"o}be, Heini, Kredens, Mayerl, P\k{e}zik, Potthast,
  Rangel, Rosso, Stamatatos, Stein, Wiegmann, Wolska, , and
  Zangerle}]{Bevendorff:2023}
\bibinfo{author}{J.~Bevendorff}, \bibinfo{author}{I.~Borrego-Obrador},
  \bibinfo{author}{M.~Chinea-R{\'i}os}, \bibinfo{author}{M.~Franco-Salvador},
  \bibinfo{author}{M.~Fr{\"o}be}, \bibinfo{author}{A.~Heini},
  \bibinfo{author}{K.~Kredens}, \bibinfo{author}{M.~Mayerl},
  \bibinfo{author}{P.~P\k{e}zik}, \bibinfo{author}{M.~Potthast},
  \bibinfo{author}{F.~Rangel}, \bibinfo{author}{P.~Rosso},
  \bibinfo{author}{E.~Stamatatos}, \bibinfo{author}{B.~Stein},
  \bibinfo{author}{M.~Wiegmann}, \bibinfo{author}{M.~Wolska}, ,
  \bibinfo{author}{E.~Zangerle},
\newblock \bibinfo{title}{{Overview of PAN 2023: Authorship Verification,
  Multi-Author Writing Style Analysis, Profiling Cryptocurrency Influencers,
  and Trigger Detection}},
\newblock in: \bibinfo{editor}{A.~Arampatzis}, \bibinfo{editor}{E.~Kanoulas},
  \bibinfo{editor}{T.~Tsikrika}, \bibinfo{editor}{A.~G. Stefanos~Vrochidis},
  \bibinfo{editor}{D.~Li}, \bibinfo{editor}{M.~Aliannejadi},
  \bibinfo{editor}{M.~Vlachos}, \bibinfo{editor}{G.~Faggioli},
  \bibinfo{editor}{N.~Ferro} (Eds.), \bibinfo{booktitle}{Experimental IR Meets
  Multilinguality, Multimodality, and Interaction. Proceedings of the
  Fourteenth International Conference of the CLEF Association (CLEF 2023)},
  Lecture Notes in Computer Science, \bibinfo{publisher}{Springer},
  \bibinfo{year}{2023}.
\bibitem[{Wiegmann et~al.(2023)Wiegmann, Wolska, Potthast, and
  Stein}]{Wiegmann:2023b}
\bibinfo{author}{M.~Wiegmann}, \bibinfo{author}{M.~Wolska},
  \bibinfo{author}{M.~Potthast}, \bibinfo{author}{B.~Stein},
\newblock \bibinfo{title}{{Overview of the Trigger Detection Task at PAN
  2023}},
\newblock in: \bibinfo{editor}{M.~Aliannejadi}, \bibinfo{editor}{G.~Faggioli},
  \bibinfo{editor}{N.~Ferro}, \bibinfo{editor}{M.~Vlachos} (Eds.),
  \bibinfo{booktitle}{Working Notes of CLEF 2023 - Conference and Labs of the
  Evaluation Forum}, \bibinfo{publisher}{CEUR-WS}, \bibinfo{year}{2023}.
\bibitem[{Wolska et~al.(2022)Wolska, Schr{\"o}der, Borchardt, Stein, and
  Potthast}]{Wolska:2022}
\bibinfo{author}{M.~Wolska}, \bibinfo{author}{C.~Schr{\"o}der},
  \bibinfo{author}{O.~Borchardt}, \bibinfo{author}{B.~Stein},
  \bibinfo{author}{M.~Potthast},
\newblock \bibinfo{title}{Trigger warnings: Bootstrapping a violence detector
  for fanfiction},
\newblock \bibinfo{journal}{arXiv preprint arXiv:2209.04409}
  (\bibinfo{year}{2022}).
\bibitem[{Wiegmann et~al.(2023)Wiegmann, Wolska, Schr{\"{o}}der, Borchardt,
  Stein, and Potthast}]{Wiegmann:2023}
\bibinfo{author}{M.~Wiegmann}, \bibinfo{author}{M.~Wolska},
  \bibinfo{author}{C.~Schr{\"{o}}der}, \bibinfo{author}{O.~Borchardt},
  \bibinfo{author}{B.~Stein}, \bibinfo{author}{M.~Potthast},
\newblock \bibinfo{title}{{Trigger Warning Assignment as a Multi-Label Document
  Classification Problem}},
\newblock in: \bibinfo{booktitle}{Proceedings of the 61th Annual Meeting of the
  Association for Computational Linguistics (Volume 1: Long Papers)},
  \bibinfo{publisher}{Association for Computational Linguistics},
  \bibinfo{address}{Toronto, Canada}, \bibinfo{year}{2023}.
\bibitem[{Devlin et~al.(2018)Devlin, Chang, Lee, and Toutanova}]{Devlin:2018}
\bibinfo{author}{J.~Devlin}, \bibinfo{author}{M.-W. Chang},
  \bibinfo{author}{K.~Lee}, \bibinfo{author}{K.~Toutanova},
\newblock \bibinfo{title}{Bert: Pre-training of deep bidirectional transformers
  for language understanding},
\newblock \bibinfo{journal}{arXiv preprint arXiv:1810.04805}
  (\bibinfo{year}{2018}).
\bibitem[{Ozcelik and Toraman(2022)}]{Ozcelik:22}
\bibinfo{author}{O.~Ozcelik}, \bibinfo{author}{C.~Toraman},
\newblock \bibinfo{title}{Named entity recognition in {T}urkish: A comparative
  study with detailed error analysis},
\newblock \bibinfo{journal}{Information Processing and Management}
  \bibinfo{volume}{59} (\bibinfo{year}{2022}) \bibinfo{pages}{103065}.
  \DOIprefix\doi{https://doi.org/10.1016/j.ipm.2022.103065}.
\bibitem[{Bird et~al.(2009)Bird, Klein, and Loper}]{Bird:2009}
\bibinfo{author}{S.~Bird}, \bibinfo{author}{E.~Klein},
  \bibinfo{author}{E.~Loper}, \bibinfo{title}{Natural Language Processing with
  Python: Analyzing Text with the Natural Language Toolkit},
  \bibinfo{publisher}{" O'Reilly Media, Inc."}, \bibinfo{year}{2009}.
\bibitem[{Liu et~al.(2019)Liu, Ott, Goyal, Du, Joshi, Chen, Levy, Lewis,
  Zettlemoyer, and Stoyanov}]{Liu:19}
\bibinfo{author}{Y.~Liu}, \bibinfo{author}{M.~Ott}, \bibinfo{author}{N.~Goyal},
  \bibinfo{author}{J.~Du}, \bibinfo{author}{M.~Joshi},
  \bibinfo{author}{D.~Chen}, \bibinfo{author}{O.~Levy},
  \bibinfo{author}{M.~Lewis}, \bibinfo{author}{L.~Zettlemoyer},
  \bibinfo{author}{V.~Stoyanov},
\newblock \bibinfo{title}{Roberta: {A} robustly optimized {BERT} pretraining
  approach},
\newblock \bibinfo{journal}{CoRR} \bibinfo{volume}{abs/1907.11692}
  (\bibinfo{year}{2019}). \href{http://arxiv.org/abs/1907.11692}{{\tt
  arXiv:1907.11692}}.
\bibitem[{Paszke et~al.(2017)Paszke, Gross, Chintala, Chanan, Yang, DeVito,
  Lin, Desmaison, Antiga, and Lerer}]{Paszke:17}
\bibinfo{author}{A.~Paszke}, \bibinfo{author}{S.~Gross},
  \bibinfo{author}{S.~Chintala}, \bibinfo{author}{G.~Chanan},
  \bibinfo{author}{E.~Yang}, \bibinfo{author}{Z.~DeVito},
  \bibinfo{author}{Z.~Lin}, \bibinfo{author}{A.~Desmaison},
  \bibinfo{author}{L.~Antiga}, \bibinfo{author}{A.~Lerer},
\newblock \bibinfo{title}{Automatic differentiation in pytorch},
\newblock in: \bibinfo{booktitle}{NIPS-W}, \bibinfo{year}{2017}.
\bibitem[{Agarap(2018)}]{Agarap:18}
\bibinfo{author}{A.~F. Agarap},
\newblock \bibinfo{title}{Deep learning using rectified linear units (relu)},
\newblock \bibinfo{journal}{CoRR} \bibinfo{volume}{abs/1803.08375}
  (\bibinfo{year}{2018}). \href{http://arxiv.org/abs/1803.08375}{{\tt
  arXiv:1803.08375}}.
\bibitem[{Buda et~al.(2018)Buda, Maki, and Mazurowski}]{Buda:18}
\bibinfo{author}{M.~Buda}, \bibinfo{author}{A.~Maki}, \bibinfo{author}{M.~A.
  Mazurowski},
\newblock \bibinfo{title}{A systematic study of the class imbalance problem in
  convolutional neural networks},
\newblock \bibinfo{journal}{Neural networks} \bibinfo{volume}{106}
  (\bibinfo{year}{2018}) \bibinfo{pages}{249--259}.
\bibitem[{Chen and Guestrin(2016)}]{Chen:2016}
\bibinfo{author}{T.~Chen}, \bibinfo{author}{C.~Guestrin},
\newblock \bibinfo{title}{Xgboost: A scalable tree boosting system},
\newblock in: \bibinfo{booktitle}{Proceedings of the 22nd ACM SIGKDD
  International Conference on Knowledge Discovery and Data Mining},
  \bibinfo{year}{2016}, pp. \bibinfo{pages}{785--794}.
\bibitem[{Salton and McGill(1984)}]{Salton:84}
\bibinfo{author}{G.~Salton}, \bibinfo{author}{M.~McGill},
  \bibinfo{title}{Introduction to Modern Information Retrieval},
  \bibinfo{publisher}{McGraw-Hill Book Company}, \bibinfo{year}{1984}.
\bibitem[{Fr{\"o}be et~al.(2023)Fr{\"o}be, Wiegmann, Kolyada, Grahm, Elstner,
  Loebe, Hagen, Stein, and Potthast}]{Froebe:2023}
\bibinfo{author}{M.~Fr{\"o}be}, \bibinfo{author}{M.~Wiegmann},
  \bibinfo{author}{N.~Kolyada}, \bibinfo{author}{B.~Grahm},
  \bibinfo{author}{T.~Elstner}, \bibinfo{author}{F.~Loebe},
  \bibinfo{author}{M.~Hagen}, \bibinfo{author}{B.~Stein},
  \bibinfo{author}{M.~Potthast},
\newblock \bibinfo{title}{{Continuous Integration for Reproducible Shared Tasks
  with TIRA.io}},
\newblock in: \bibinfo{editor}{J.~Kamps}, \bibinfo{editor}{L.~Goeuriot},
  \bibinfo{editor}{F.~Crestani}, \bibinfo{editor}{M.~Maistro},
  \bibinfo{editor}{H.~Joho}, \bibinfo{editor}{B.~Davis},
  \bibinfo{editor}{C.~Gurrin}, \bibinfo{editor}{U.~Kruschwitz},
  \bibinfo{editor}{A.~Caputo} (Eds.), \bibinfo{booktitle}{Advances in
  Information Retrieval. 45th European Conference on {IR} Research ({ECIR}
  2023)}, Lecture Notes in Computer Science, \bibinfo{publisher}{Springer},
  \bibinfo{address}{Berlin Heidelberg New York}, \bibinfo{year}{2023}, pp.
  \bibinfo{pages}{236--241}.

\end{thebibliography}

\end{document}